\documentclass[aps,pra,superscriptaddress,twocolumn,longbibliography]{revtex4-1}
\usepackage{mathrsfs}
\usepackage{amsfonts}
\usepackage{amssymb}
\usepackage{amsmath}
\usepackage{graphicx}
\usepackage{sidecap}
\usepackage{color}
\usepackage[colorlinks=true, urlcolor=blue, citecolor=blue,linkcolor=blue,citebordercolor={1 0 0},linkbordercolor={0 0 1}]{hyperref}
\usepackage{subeqnarray}
\usepackage{euscript} 
\usepackage{hyperref}

\usepackage[version=4]{mhchem}
\usepackage{multirow}
\usepackage[ruled,vlined]{algorithm2e}
\usepackage[]{algpseudocode}
\usepackage{shortcuts}
\usepackage{bm}
\usepackage{color}
\usepackage{amsthm,parskip,setspace,tabularx, wrapfig,float,enumerate} 
\usepackage{tikz, setspace, subfig, grffile}
\usepackage{subfig}
\usepackage{paralist}
\usetikzlibrary{bayesnet}
\usepackage{float}

\renewcommand{\eqref}[1]{Eq.~(\ref{#1})} 
\usepackage[columnwise]{lineno} 

\newcommand{\utchem}{Department  of  Chemistry,  University  of  Toronto,  Toronto,  Ontario  M5G 1Z8,  Canada}
\newcommand{\harvardchem}{Department of Chemistry and Chemical Biology, Harvard University, Cambridge, Massachusetts 02138, USA}
\newcommand{\harvardcs}{Department of Computer Science, Harvard University, Cambridge, Massachusetts 02138, USA}
\newcommand{\utcomp}{Department  of  Computer Science,  University  of  Toronto,  Toronto,  Ontario  M5S 2E4,  Canada}
\newcommand{\vectorinst}{Vector  Institute  for  Artificial  Intelligence,  Toronto,  Ontario  M5S  1M1,  Canada}
\newcommand{\cifar}{Canadian  Institute  for  Advanced  Research,  Toronto,  Ontario  M5G  1Z8,  Canada}

\begin{document}

\title{Scalable Fragment-Based 3D Molecular Design with Reinforcement Learning}

\author{Daniel Flam-Shepherd}
\thanks{These authors contributed equally. \urlstyle{same} }
\affiliation{\utcomp}
\affiliation{\vectorinst}

\author{Alexander Zhigalin}
\thanks{These authors contributed equally. \urlstyle{same} }
\affiliation{\harvardchem}
\affiliation{\harvardcs}

\author{Al\'an Aspuru-Guzik}
\affiliation{\utcomp}
\affiliation{\vectorinst}
\affiliation{\utchem}
\affiliation{\cifar}

\begin{abstract}
\noindent Machine learning has the potential to automate molecular design and drastically accelerate the discovery of new functional compounds.
Towards this goal, generative models and reinforcement learning (RL) using string and graph representations have been successfully used to search for novel molecules.  
However, these approaches are limited since their representations ignore the three-dimensional (3D) structure of molecules. 
In fact, geometry plays an important role in many applications in inverse molecular design, especially in drug discovery. 
Thus, it is important to build models that can generate molecular structures in 3D space based on property-oriented geometric constraints. To address this, one approach is to generate molecules as 3D point clouds by sequentially placing atoms at locations in space -- this allows the process to be guided by physical quantities such as energy or other properties. However, this approach is inefficient as placing individual atoms makes the exploration unnecessarily deep, limiting the complexity of molecules that can be generated. Moreover, when optimizing a molecule, organic and medicinal chemists use known fragments and functional groups, not single atoms. We introduce a novel RL framework for scalable 3D design that uses a hierarchical agent to build molecules by placing molecular substructures sequentially in 3D space, thus attempting to build on the existing human knowledge in the field of molecular design. In a variety of experiments with different substructures, we show that our agent, guided only by energy considerations, can efficiently learn to produce molecules with over 100 atoms from many distributions including drug-like molecules, organic LED molecules, and biomolecules.
\end{abstract}

\maketitle


\section{Introduction}

Chemical space is massive -- the number of drug-like molecules alone is estimated to be as large as $10^{60}$ \citep{Polishchuk2013}. For larger molecules, the size of the space grows considerably. This makes discovering novel functional compounds for drug design and materials engineering an immensely challenging task \citep{Schneider2019}. This is further complicated by the fact that the 3D structure of a molecule is an important determinant of molecular properties and, as a result, is highly relevant to molecular design. However, molecular geometry is rarely taken into consideration in current machine learning (ML) approaches aimed at accelerating the search through chemical space. For example, generative models of molecules are often trained on SMILES \cite{Weininger1988} string representations \citep{Gomez-Bombarelli2016, Segler2017b, Kusner2017, Dai2018} or trained directly on molecular graphs \cite{Liu2018,kwon2019efficient,simonovsky2018graphvae, flam2021mpgvae, seff2019discrete}. Similarly, SMILES strings and graphs have been successfully used in reinforcement learning \cite{guimaraes2017objective,You2018}. In contrast, supervised models \cite{gilmer2017neural, schutt2018schnet, pattanaik2020message,flam2021neural, klicpera2021gemnet} made early use of molecular geometry -- mostly as input features for property prediction. 

The SMILES strings and molecular graphs used in recent ML approaches are simplified approximations of the physical, three dimensional structure of molecules that exist in nature. Additionally, these representations limit the class of molecules that can be generated, as some compounds such as metal-organic frameworks cannot be described at all without the 3D information of their atoms. Furthermore, using string and graph representations, it is not possible to generate molecules with geometric constraints on the arrangement of atoms.

\begin{figure}[t]
\centering
\includegraphics[width=0.99\columnwidth]{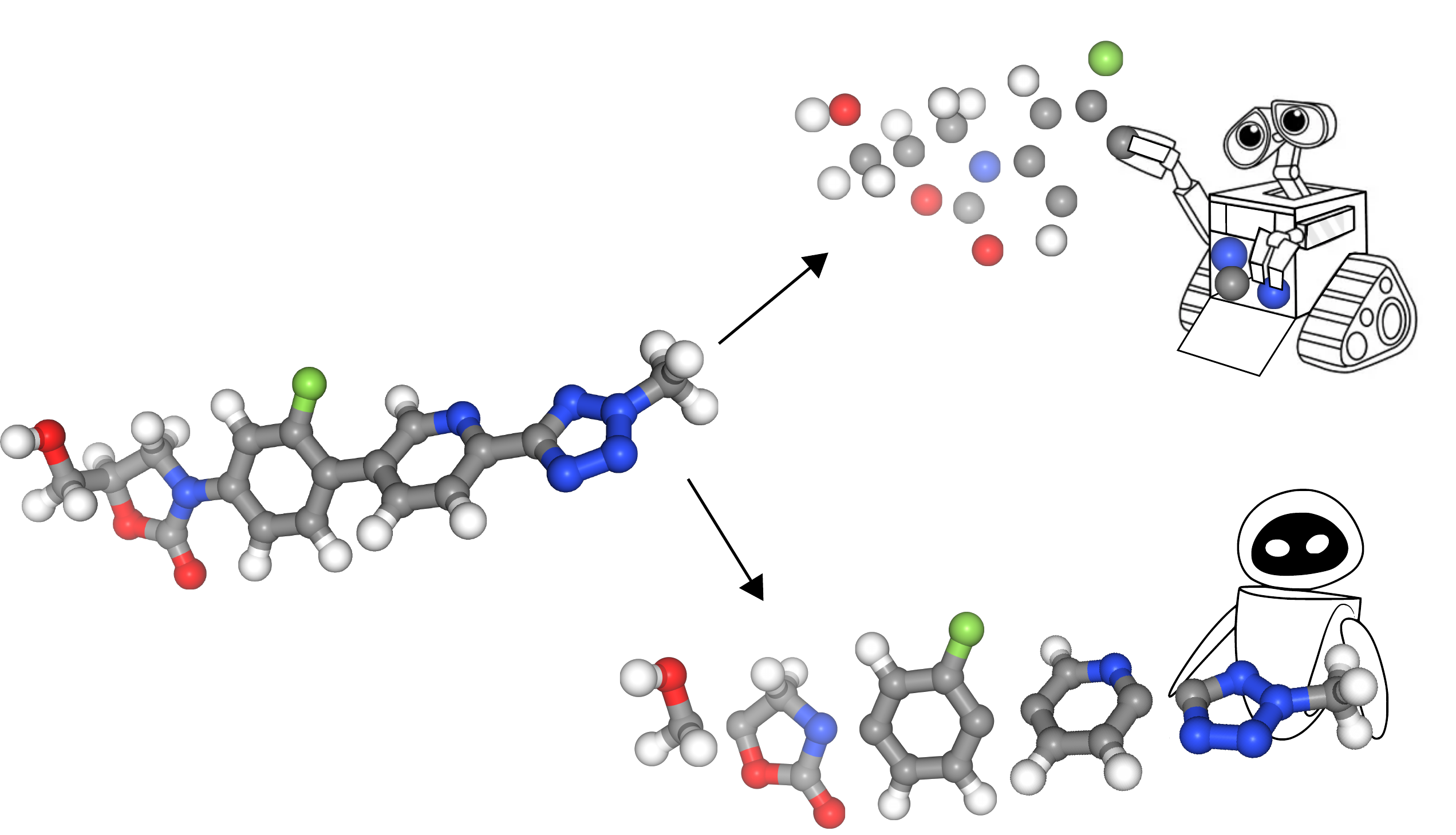}
\caption{\raggedright
Two RL agents -- WALL-E and EVE \cite{walle2015} -- have to design the antibiotic Tedizolid (left).
WALL-E places single atoms in order to complete the task (top). In contrast (bottom), EVE uses our framework and places whole fragments (visualized by rendering bonds within -- but not between -- individual fragments).}
\label{fig:intro}
\end{figure}

A few recent approaches do take 3D structure into consideration. For example, G-SchNet \cite{Gebauer2019} is an auto-regressive generative model using SchNet \cite{Schutt2017} that builds molecules as point clouds of atoms. The model sequentially places atoms in Cartesian coordinates using a 3D grid of interatomic distances. G-SchNet has demonstrated limited scalability only generating small organic molecules from QM9 \cite{Ramakrishnan2014}. In addition, \cite{simm2020reinforcement} proposed a novel RL framework MolGym for 3D molecular design, where an agent places atoms from a given bag of atoms onto a 3D canvas using internal coordinates (as visualized in Fig. \ref{fig:intro}, top). MolGym uses a reward function based on fundamental physical properties such as energy and thus it is not restricted to generating a specific type of molecule. This approach was further refined in \cite{simm2020symmetry} through the use of a rotationally covariant state-action representation based on a series expansion of spherical harmonics. This allowed them to generate structures from a slightly larger class of symmetric molecules with more complex geometry. Regardless, MolGym also has also only been able to design small QM9 sized molecules.

Designing molecules in 3D space by placing individual atoms makes it necessary for the model to learn to build substructures from scratch. However, scientific approaches to molecular design use previously discovered substructures, for example fragment-based drug design \cite{murray2009rise, over2013natural} or high-throughput virtual screening where libraries are combinatorially built from a set of known fragments \cite{shoichet2004virtual,walters1998virtual, gomez2016design}. This has motivated many machine learning approaches to use fragments for graph-based molecular design 
\cite{jin2018junction, jin2018learning, jin2020multi, jin2020hierarchical}.  

In addition, for models that place single atoms, there will be considerable time spent explicitly placing individual hydrogen atoms in order to design valid molecules.  
Furthermore, to design important geometric substructures, the model will be forced to generate intermediaries that are either chemically invalid or do not make physical sense.
For this reason, when the generative process is guided by energy, it will be very unlikely that the model learns to build important geometric substructures. This is why MolGym \cite{simm2020reinforcement, simm2020symmetry} has only been able to demonstrate the design of small molecules ($<30$ atoms including hydrogen)) with simple substructures.

In this work, we introduce a scalable architecture for 3D fragment-based molecular design with reinforcement learning. Our approach uses a hierarchical agent to sequentially place any molecular structure in 3D space -- single atoms, molecular fragments, and, as we show, even entire molecules. Endowed with this prior, the policy can more efficiently learn to produce molecules that are larger and more complex. Furthermore, by using some specific substructures, the design process can be tailored to a particular application \cite{Gomez-Bombarelli2016, over2013natural}. Our approach is visualized in Figure \ref{fig:intro} (bottom), carried out by the RL agent EVE.

We experiment with our framework in a fixed space, focusing on the general task of designing valid molecules in 3D using a variety of substructures.
For that purpose, we devise a general energy based reward that accounts for our fragment-based generative process. We demonstrate the scalability of our approach by building different kinds of molecules with over 100 atoms including drug-like molecules, organic LED molecules, and biomolecules.

\section{Reinforcement learning framework for fragment-based 3D molecular design}

\subsection{Problem Definition}

We represent a molecule $\mathcal{M}=\{ (\B x_i, \bm r_i )\}_{i=1}^n$ as a point cloud of $n$ atoms with features $\B x_i \in \mathbb{R}^{n_a}$ and coordinates (positions) $\bm r_i \in \mathbb{R}^3$ where $n_a$ is number of unique atoms possible. We work with the set of atoms including \{C,H,N,O,F,S\}. The design task we focus on is to construct a molecule in 3D space using every fragment in a fixed multiset $\mathcal{F}$ of $m$ fragment, given by $\mathcal{F} = \{(\B X_{1}, \bm R_{1}),\dots , (\B X_{f}, \bm R_{f}),\dots, (\B X_{m}, \bm R_{m})\}$.  
Each fragment $f$ is a small point cloud with $1 <n_f<n$ atoms with its own initial positions $\bm R_f = (\bm r_{\footnotesize 1}, \dots , \bm r_{n_{\scriptscriptstyle f}}) \in \mathbb{R}^{3 \times n_f }$ 
and atom features $ \B X_f =( \B x_{\footnotesize 1} , \dots , \B x_{n_{\scriptscriptstyle f}} )\in \mathbb{R}^{n_a \times n_f}$.  

The overall objective is to generate valid molecules in 3D space guided by the energy $ \mathrm{E}(\mathcal{M})$. 
After placing all the fragments from $\mathcal{F}$ in space, the final molecule $\mathcal{M}$ will be a combination of all the fragments in the multiset. 
Effectively, the design problem can be distilled into a inference problem solved by specifying 1) which hydrogens should be removed from the fragment point clouds in order to anchor fragments together before placement and 2) What should the final coordinates of the remaining atoms in the fragment point cloud after placement. 

The initial positions of the atoms in each fragment define an implicit prior on geometry for the molecule
defined by the prior geometry $\bm R_f $ for each structure in the fragment multiset $ \mathcal{F}$. 
For this prior, we can use predefined conformer generation from RDKit \cite{RDKit2019093} or OpenBabel \cite{o2011open} as well as recent machine learning approaches for conditional conformation generation \cite{Simm2020GraphDG,ganea2021geomol, gogineni2020torsionnet}.

\subsection{Fragment-Based Generation as a MDP}\label{sec:setup}

We assume a standard RL framework where an agent interacts with a fully observable environment following deterministic dynamics in order to maximize some reward. We design a sequential molecular design process as a point cloud generation task using multisets of molecular fragments that are treated as small point clouds.

We formulate it as a Markov decision process (MDP) $(\mathcal{S}, \mathcal{A}, \mathcal{T}, \gamma, T, r)$.
The state space $\mathcal{S}$ is the set of states consisting of all final and intermediate molecules. 
The action space $\mathcal{A}$ is the set of actions that describe the addition and orientation of a new fragment to the current molecule at each step. 
The transition function $\mathcal{T}: \mathcal{S} \times \mathcal{A} \mapsto S$ determines the outcome of carrying out an action (and can incorporate any geometric constraints). Also we define a discount factor $\gamma \in (0, 1]$, time horizon $T$, 
and a reward function $r: \mathcal{S} \times \mathcal{A} \mapsto \mathbb{R}$ which quantifies how the action $a_t$ impacts the molecule.

\begin{figure*}[t]
\centering
\begin{tikzpicture}
    \draw (0, 0) node[inner sep=0] {\includegraphics[width=0.725\textwidth]{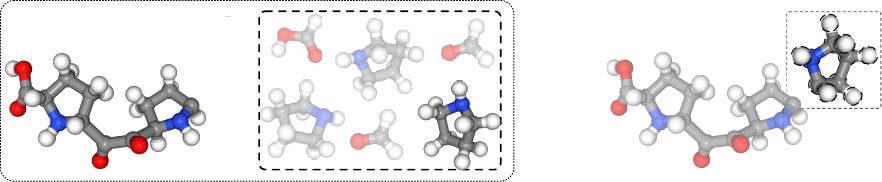}};
    \draw (-6.58, 1.3) node[] {\textbf{a}};
    \draw (-5.72, 1.12) node[] {state $s_t$};
    \draw (-4.5, 0.67) node[] {$\mathcal{M}_t$};
    \draw (-2.4, 0.25) node[] {$\mathcal{F}_t$};
    \draw (4.34, 0.62) node[] {$\mathcal{M}_{t+1}$};
    \draw (5.34, 0.95) node[] {$a_t$};
    \draw (2.7, 1.2) node[] {action};
     \draw (6.52, -0.63) node[] {$(\B X_t, \bm R_t)$};
\end{tikzpicture}

\vspace{.25cm}

\begin{tikzpicture}
    \draw (0, 0) node[inner sep=0] {\includegraphics[width=0.95\textwidth]{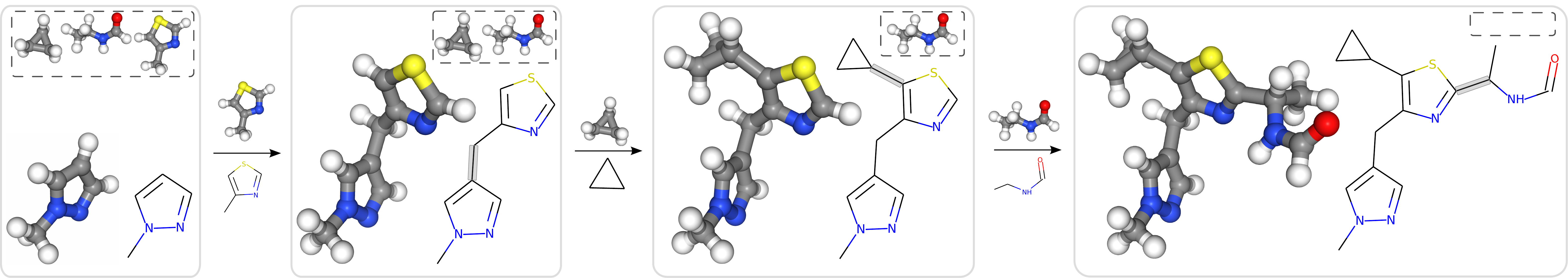}};
    \draw (-8.5, 1.55) node[] {\textbf{b}};
\end{tikzpicture}

\caption{\raggedright
\textbf{RL Basics.} \textbf{a.} The state $s_t$ is defined by the molecule $\mathcal{M}_t$ at step $t$ (left) and the current condition of the fragment multiset $\mathcal{F}_t$ (center, faded fragments have already been placed on the molecule). A single action $a_t$ (right) consists of expanding the current molecule to $\mathcal{M}_{t+1}$ by placing a fragment or an entire molecule as a point cloud of atoms $\B X_t $ with added positions $\bm R_t$. \textbf{b.} The rollout of an episode with the fragment multiset $\mathcal{F} = \{\ce{C4H6N2},\ce{C4H5NS},\ce{C3H6},\ce{C3H7NO} \}$.
The agent constructs a molecule with the formula $\ce{C14H22N4OS}$, starting with the substructure $\ce{C4H6N2}$ by sequentially placing fragments from the multiset without replacement.}
\label{fig:rl}
\end{figure*}
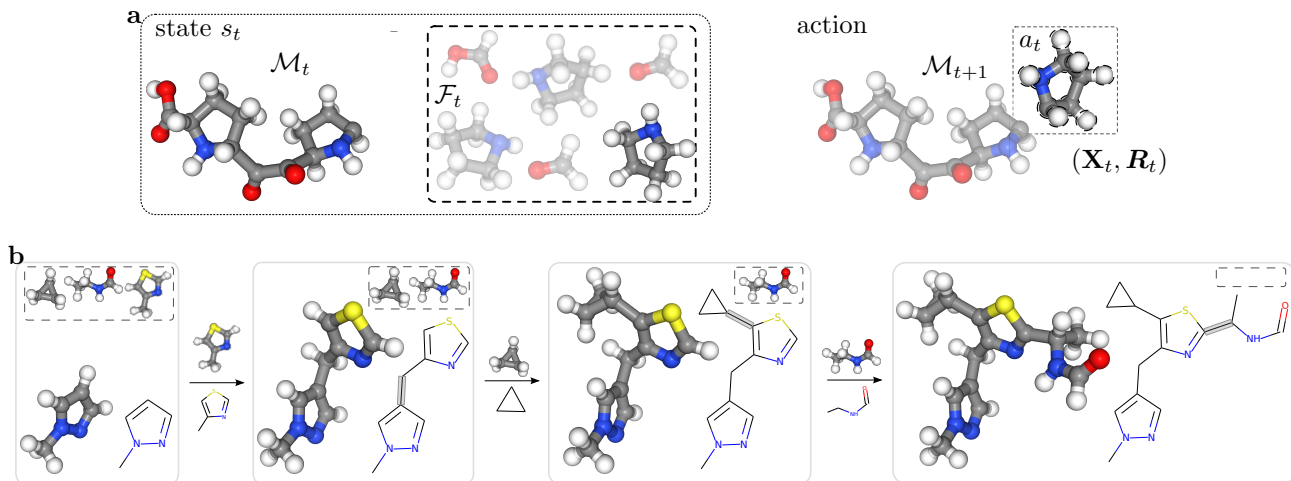

\subsection{State and Action Representation.} 

\textbf{State.} We define the environment state $s_t$ as the intermediate molecule $\mathcal{M}_t$ that has been generated so far at step $t$, as well as the condition of the structure or fragment multiset $\mathcal{F}_t$. 
The current molecule can be represented as a point cloud of all structures placed up to the current step $t$: 
$ \mathcal{M}_t = \left  \{ (\B X_{0}, \bm R_{0}) , \dots, (\B X_{t-1}, \bm R_{t-1})\right \} $
where $(\B X_{\ell}, \bm R_{\ell})$ are fragment point clouds removed from the structures multiset with some hydrogens removed during fragment placement at each previous step. The current condition of the multiset (fragments that have not been placed yet) is also used: $\mathcal{F}_t = \left\{ f \ | \ (\B X_{f}, \bm R_{f}) \notin \mathcal{M}_{t} \right\}$. The state representation is shown in Figure \ref{fig:rl}\textbf{a}.  
The agent starts from an initial fragment $(\B X_{0}, \bm R_{0})$ which can be selected beforehand or chosen at random from the multiset.
The agent can either start building by placing the first fragment at the origin or by extending a given molecule. 

\textbf{Action.} In our framework, we use a high level action -- we can place any point cloud of atoms at every step, including single atoms, fragments, or entire molecules. Every action expands the generated molecule $\mathcal{M}_{t}$ by adding to it a molecular substructure $ a_t = (\mathbf{X}_t, \bm R_t) $ which has been taken from the fragment multiset and had a single hydrogen removed, whose location is used to anchor the fragment to the molecule. However, we do not directly place every individual atom in the point cloud. Rather, the agent learns a hierarchy of sub-actions to find where and how to place an additional point cloud  (Figure \ref{fig:rl}\textbf{a}).

Figure \ref{fig:rl}\textbf{b} shows the rollout of an episode. Initially, a fragment is placed at the origin. Fragments are then sequentially drawn from a multiset and placed in space until no fragments remain. 

\subsection{Reward Function.} 

As we are designing molecules using Cartesian coordinates, we can directly use energy to guide our agent. Therefore, we can construct a reward function that allows us to design molecules with lower energy $\mathrm{E} \in \mathbb{R}$. To do so, for the action $a_t$ taken in the state $s_t$ at every step, we formulate the reward function $r(s_t, a_t)$ as:
\begin{align}\label{eq:energy}
    r(s_t, a_t) = -\left [\mathrm{E}(\mathcal{M}_{t+1}) - \mathrm{E}(\mathcal{M}_{t}^{+f}) \right ] 
\end{align}
where $\mathrm{E}(\mathcal{M}_{t+1})$ is the energy of the molecule after the action $a_t$ is taken and 
\begin{align}\label{eq:frag}
    \mathrm{E}(\mathcal{M}_{t}^{+f}) = \mathrm{E}(\mathcal{M}_{t})+\mathrm{E}(  \{ (\B X_{t}, \bm R'_{t})\} ) 
\end{align}
where $\mathrm{E}(\mathcal{M}_{t}) $ is the energy of the current molecule at step $t$ 
and $\mathrm{E}(  \{ (\B X_{t}, \bm R'_{t})\} $ is the energy of the added fragment 
under the prior geometry $\bm R'_t$ where hydrogens were removed in both the molecule and the fragment to form a covalent bond in $\mathcal{M}_{t+1}$.

The reward is essentially the negative difference in energy between the resulting molecule and the two non-interacting building blocks that form the new covalent bond. The agent is rewarded for making the most stable covalent bonds and orienting the fragments in place so that the most stable conformation is generated. To find energy $E(\mathcal{M})$, we use a fast semi-empirical method (PM6) \citep{Stewart2007} implemented in \textsc{Sparrow} \citep{Husch2018a,Bosia2019}.

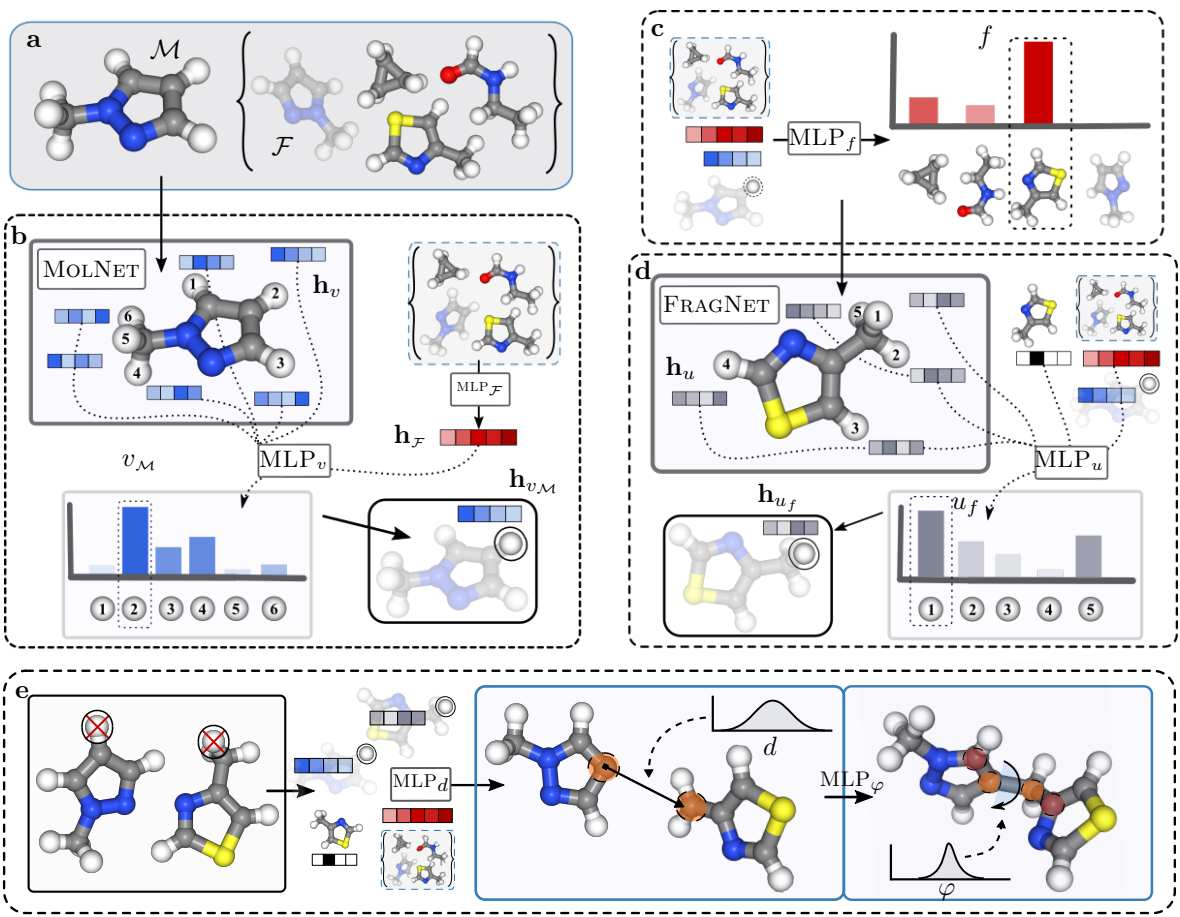
\begin{figure*}[t]
\centering
\begin{tikzpicture}
    \draw (0, 0) node[inner sep=0] {\includegraphics[width=0.87\textwidth]{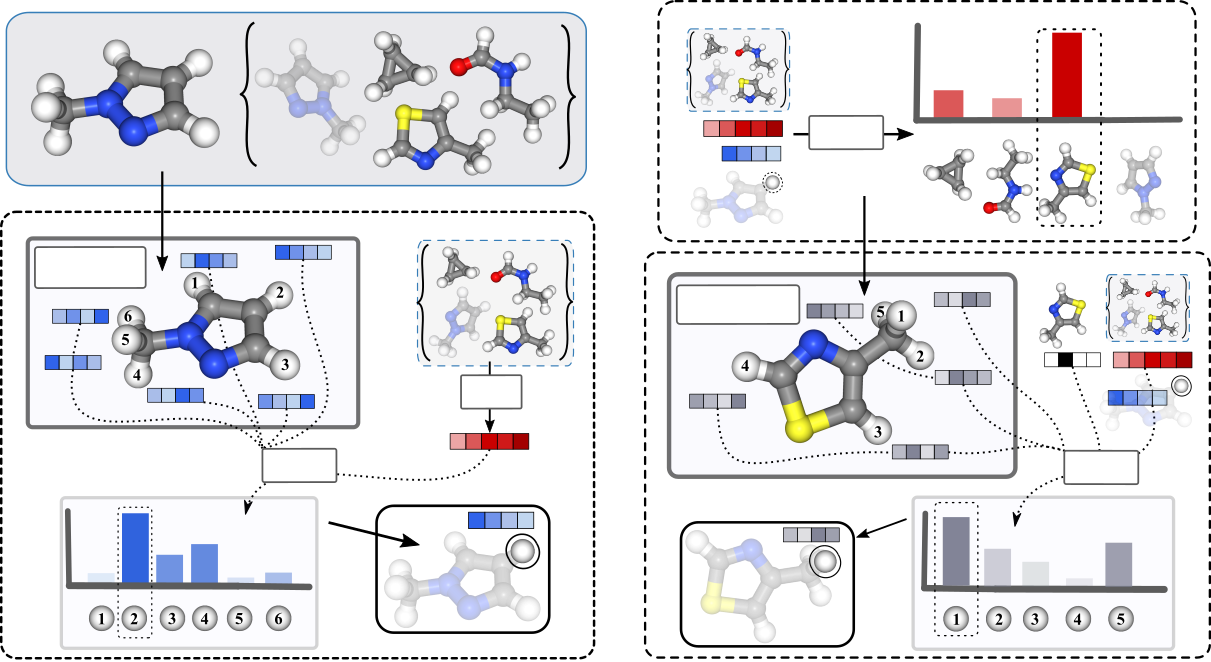}};
    \draw (-7.4, 3.85) node[] {\textbf{a}};
    \draw (-5.65, 3.75) node[] {$\mathcal{M}$};
    \draw (-4.1, 2.45) node[] {$\mathcal{F}$};
    \draw (-7.6, 1.27) node[] {\textbf{b}};
    \draw (-6.0, -1.75) node[]{$v_{\scriptscriptstyle\mathcal{M}} $};
    \draw (-2.4, -1.4) node[]{$\B h_{\scriptscriptstyle\mathcal{F}}$};
    \draw (-3.5, 0.6) node[]{$\B h_v$};
    \draw (-0.75, -2.0) node[]{$\B h_{v_{\scriptscriptstyle\mathcal{M}}}$};
    \draw (2.5, -2.225) node[]{$\B h_{u_f}$};
    \draw (0.89, 4.0) node[] {\textbf{c}};
    \draw (5.25, 3.9) node[] {$f$};
    \draw (0.7, 0.81) node[] {\textbf{d}};
    \draw (-3.95, -1.72) node[] {\small$\textsc{MLP}_v$};
    \draw (-1.465, -0.775) node[] {$\scriptscriptstyle\textsc{MLP}_{\tiny\mathcal{\tiny F}}$};
    \draw (3.09, 2.53) node[] {\small$\textsc{MLP}_f$};
    \draw (6.36, -1.725) node[] {\small$\textsc{MLP}_u$};    
    \draw (-6.64, 0.825) node[] {\small \textsc{MolNet}};
    \draw (1.675, 0.35) node[] {\small \textsc{FragNet}};
    \draw (1.2, -0.5) node[]{$\B h_u$};
    \draw (4.99, -2.35) node[] {$u_f$};
\end{tikzpicture}

\begin{tikzpicture}
    \draw (0, 0) node[inner sep=0] {\includegraphics[width=0.87\textwidth]{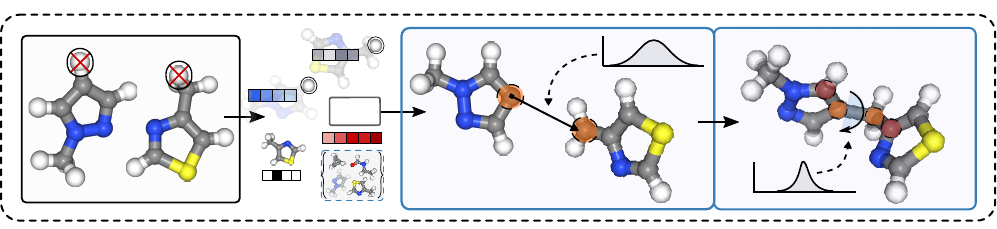}};
    \draw (-7.54, 1.28) node[] {\textbf{e}};
    \draw (-2.24, 0.067) node[] { $\scriptsize \textsc{MLP}_{d}$};
    \draw (3.5, 0.067) node[] { $\footnotesize \textsc{MLP}_{\varphi}$};
    \draw (2.4, 0.56) node[] {$d$};
    \draw (4.75, -1.375) node[] {$\varphi$};
\end{tikzpicture}
\caption{\raggedright
\textbf{The Policy Network.} \textbf{a.} Molecule and the remaining fragments in the multiset. \textbf{b.} Using the representation of atoms in the molecule and fragments in the set, we select a hydrogen on the molecule. \textbf{c.} Using the representations of the selected hydrogen and the fragment set, we select a new fragment to add to the molecule. \textbf{d.} Using the fragment atomic embeddings and the representation of the multiset, we select a hydrogen on the fragment to anchor to the molecule. \textbf{e.} Selected hydrogens are removed and two MLPs produce the distance and rotation angle between fragments using the embeddings of the selected hydrogens and the representations of the fragment multiset and the chosen fragment. The added fragment is positioned in 3D space next to the molecule.}
\label{fig:model}
\end{figure*}

\section{The Policy Network}

The use of substructures with prior geometry makes the molecular design process substantially more complicated. 
For the agent to learn how to place the fragments in space to build a molecule, 
three main questions must be resolved: 
1) Where on the generated molecule should the fragment be added? 2) In what order should the fragments be added? 3) How should the fragment be placed on the molecule?  

In order to answer these questions, we introduce two main networks in our actor: 
one to learn embeddings on the molecule and 
one to learn embeddings on the fragment being added at every step. These representations are used by the agent to perform sub-actions that answer these three main questions. Specifically, the action is broken into a set of discrete sub-actions for atom and fragment selections modeled by categorical distributions and continuous sub-actions for geometry modeled by Gaussian distributions. 

First, the agent selects a hydrogen atom $v_{\scriptscriptstyle\mathcal{M}}$ on the generated molecule then conditionally chooses a fragment $f$ and a hydrogen on the fragment $u_f$ to anchor the placement of the fragment on the molecule. Given a state $s$, we assume the distribution of selections is:
\begin{align} p(u_f ,f, v_{\scriptscriptstyle\mathcal{M}}|s) =  p(u_f |f, v_{\scriptscriptstyle\mathcal{M}}, s)  p(f  | v_{\scriptscriptstyle\mathcal{M}}, s)  p(v_{\scriptscriptstyle\mathcal{M}} | s)
\end{align}
The hydrogens will be removed and the remaining atoms in the fragment are placed. Instead of placing them in Cartesian coordinates which are not invariant under translation and rotation, we model the position of the fragment in internal coordinates and then map each atom position back to Cartesian coordinates.

Figures \ref{fig:model}\textbf{a-d} show how the policy anchors the added fragment to the molecule by making a few sub-actions using the state as an input. 

\subsection{Molecule Sub-policy} 

At every step, we use a network \textsc{MolNet} to produce embeddings of the atoms in the current molecule $\mathcal{M}$
\begin{equation}\label{eq:molnet}
  \{ \B h_v\}_{v\in \mathcal{M}} = \textsc{MolNet} ( \{ \mathbf{x}_v, \bm r_v \}_{v\in \mathcal{M}} ) 
\end{equation}
We want to produce a representation of each atom $v$ in $\mathcal{M}$ using information about its local neighbourhood.
We can use any domain specific neural network for \textsc{MolNet} that is invariant under translation and rotation. Since in this work we use point cloud representations of molecules, we use continuous convolutions, namely, \textsc{SchNet} \citep{Schutt2017}.

In order to select an anchor location on the molecule to use when placing the fragment-- we need to produce a distribution over atoms. 
For this we construct feature vectors: 
$ [\B h_v \oplus \B h_{\scriptscriptstyle\mathcal{F}} ]_{v\in \mathcal{M}}$ where $\B h_{\scriptscriptstyle\mathcal{F}} = \textsc{MLP}_{\scriptscriptstyle\mathcal{F}}(\bm x_{\scriptscriptstyle\mathcal{F}})$ 
is the learned representation of the fragment multiset using a multilayer perceptron $\textsc{MLP}_{\scriptscriptstyle\mathcal{F}}$ with input 
$\bm x_{\scriptscriptstyle\mathcal{F}}=[n_f \ \text{for} \ f \in \mathcal{F}]$ 
where $n_f$ is the count of fragments $f$ in the current fragment multiset $\mathcal{F}$. 
Using a masked softmax we compute the probabilities over candidate atoms  
\begin{equation}\label{eq:molnet}
 p(v_{\scriptscriptstyle\mathcal{M}} | s) = \frac{\mathbb{M}_{v} \exp \left( \textsc{MLP}_v[\B h_v \oplus \B h_{\scriptscriptstyle\mathcal{F}} ] \right )  }{\sum_{v\in \mathcal{M}} \mathbb{M}_{v}  \exp \left( \textsc{MLP}_v[\B h_v \oplus \B h_{\scriptscriptstyle\mathcal{F}} ] \right )}  
\end{equation}
The mask $\mathbb{M}_{v} = \mathbb{I}[\mathrm{Z}_v=1]$ ensures the selection of a hydrogen atom i.e. atom with atomic number $\mathrm{Z}_v=1$. The selected hydrogen atom to be removed is then sampled $v_{\scriptscriptstyle \mathcal{M}} \sim \text{Categorical}( p(v_{\scriptscriptstyle \mathcal{M}} | s) )$ (Figure \ref{fig:model}\textbf{b}).

\subsection{Fragment Sub-policy} 
We must now anchor a fragment to be placed near the neighboring heavy atom to the selected hydrogen on the molecule. To do this, we need another network to learn representations for the fragments. We call this sub-policy network \textsc{FragNet} and it is only trained on fragment point clouds from the multiset. Instead of learning over all fragments at every step, we first conditionally select a single fragment from the multiset to place. Using the representation of the selected atom and fragment space 
$\B h_f  =  \textsc{MLP}_f([\B h_{v_{\scriptscriptstyle \mathcal{M}} }\oplus \B h_{\scriptscriptstyle\mathcal{F}}  ] )$,
we compute a distribution over possible fragment selections
\begin{equation}\label{eq:molnet}
 p(f  | \ v_{\scriptscriptstyle\mathcal{M}}, s) = {\scriptstyle \mathbb{ M} }_{f} \exp( \B h_f) \big/ \textstyle \sum_{f'\in \mathcal{F} }  {\scriptstyle \mathbb{ M} }_{f'}   \exp( \B h_{f'})
\end{equation}
masking out fragments already placed : $ {\scriptstyle \mathbb{ M} }_{f} = \mathbb{I}[n_f>0] $. From this distribution we sample a fragment to place as another sub-action. This is shown in Figure \ref{fig:model}\textbf{c}. Next, we select a fragment hydrogen atom $u$ that will be used to anchor the fragment to the molecule. Now we can use the fragment $f$ and $\textsc{FragNet}$ to get atomic embeddings: 
\begin{equation}\label{eq:fragnet}
  \{ \B h_u\}_{u\in f} = \textsc{FragNet} ( \{ \mathbf{x}_u, \bm r_u \}_{u\in f} ) 
\end{equation}
then once again we construct the feature vectors for all atoms in the fragment 
$u\in f$ :  $\B H^{f,{v_{\scriptscriptstyle\mathcal{M}}}}_u = \B h_u \oplus \B h _{v_{\scriptscriptstyle\mathcal{M}}} \oplus \B h_{\scriptscriptstyle\mathcal{F}}\oplus \bm x_f $ with the selected hydrogen embedding, fragment space representation, and fragment one-hot encoding. 
Using a masked softmax, we compute the probabilities over the candidate atoms:
\begin{equation}\label{eq:molnet}
 p(u_f |f, v_{\scriptscriptstyle\mathcal{M}}, s) = \frac{\mathbb{M}_{u} \exp \left( \textsc{MLP}_u[\B H^{f,{v_{\scriptscriptstyle\mathcal{M}}}}_u ] \right )  }{\sum_{u'\in f} \mathbb{M}_{u'}  \exp \left( \textsc{MLP}_u[\B H^{f,{v_{\scriptscriptstyle\mathcal{M}}}}_{u'}] \right )}  
\end{equation}
We mask out non-hydrogens: $\mathbb{M}_{u} = \mathbb{I}[\mathrm{Z}_u=1]$
and sample a hydrogen atom $u_f \sim \text{Categorical}( p(u_f |f, v_{\scriptscriptstyle\mathcal{M}}, s) )$
on the selected fragment. This is visualized in Figure \ref{fig:model}\textbf{d}. 
Every discrete sub-action serves some purpose to streamline the final placement of the fragment 
using the continuous sub-actions. In addition, by selecting hydrogens, it allows us to avoid keeping track of the graph and make valid placements.

\subsection{Fragment Placement and Orientation} 

Using the discrete selections, the agent can now conditionally generate the final geometry of the molecule using the prior geometry and two continuous sub-actions. The continuous sub-actions are a distance $d$ and rotation angle $\varphi$. This means that the entire policy $\pi_{\theta} (d, \varphi, u_f ,f, v_{\scriptscriptstyle\mathcal{M}} | s)$ factorizes into two terms    
$p(d, \varphi | u_f ,f, v_{\scriptscriptstyle\mathcal{M}} , s)  \times p(u_f ,f, v_{\scriptscriptstyle\mathcal{M}}| s) $
The agent assumes the distance and rotation angle are independent and $p(d, \varphi | u_f ,f, v_{\scriptscriptstyle\mathcal{M}} , s)$ 
is the product of two univariate Gaussians with standard deviations that are fixed global parameters.  
Parameter $d$ is the distance between the heavy atoms that were closest to the removed hydrogens. We use a MLP to predict the mean distance $d = ||\B r_{a_f}-\bm r_{a_{\scriptscriptstyle\mathcal{M}} }||$ where $a_{\scriptscriptstyle\mathcal{M}}$ and $a_f$ are the heavy atoms anchoring the molecule to the fragment, that is 
\begin{equation}\label{eq:dist}
    \mathbb{E}[d] = \textsc{MLP}_d([\B h_{u_f} \oplus \B h _{v_{\scriptscriptstyle\mathcal{M}}} \oplus \B h_{\scriptscriptstyle\mathcal{F}}\oplus \bm x_f  ])
\end{equation}
The agent also uses a rotation angle $\varphi \in [-\pi, \pi]$ to orient the fragment in place. 
This is a dihedral angle defined by $a_{\scriptscriptstyle\mathcal{M}}$ and $a_f$ the heavy atoms anchoring the molecule to the fragment and their two closest neighbouring atoms. 
Using the learning representations from the discrete actions we predict its mean with another MLP
\begin{equation}\label{eq:rot}
    \mathbb{E}[\varphi] = \textsc{MLP}_{\varphi}([\B h_{u_f} \oplus \B h _{v_{\scriptscriptstyle\mathcal{M}}} \oplus \B h_{\scriptscriptstyle\mathcal{F}} \oplus \bm x_f ])
\end{equation}
As shown in Fig.~\ref{fig:model}\textbf{e}, the selected hydrogens are removed and the fragment is anchored at distance $d$ from the molecule and rotated into place. Using the rotation angle and distance, we can map the fragment atom positions back to Cartesian coordinates $\{\bm r_v \in \mathbb{R}^3\}_{v\in f}$. We then model the sign of the rotation angle similar to \cite{simm2020reinforcement} (details in  \ref{app:methods}4). We also use a MLP layer to compute a value for the critic (details in  \ref{app:methods}3).  
To train the policy we use PPO \citep{schulman2017proximal} (details in  \ref{app:methods}2).

\begin{table*}[t]
\centering
\label{tab:metrics} 
\begin{tabular}{@{}cccccccccc}
& \multicolumn{3}{c|}{Drug-like} & \multicolumn{3}{|c|}{Organic LED} & \multicolumn{3}{|c}{Biomolecules} \\ 
\hline  \parbox[t]{4mm}{\multirow{2}{*}{\rotatebox[origin=c]{90}{Sets}}}
 & \scriptsize \ce{C14H22N4OS} & \scriptsize \ce{C16H27N5O2} & \scriptsize \ce{C29H31N7O} & \scriptsize \ce{C34H23N5}  &\scriptsize \ce{C49H28N4O} &\scriptsize \ce{C67H50N6}  &\scriptsize \ce{C20H38N6O4} &\scriptsize \ce{C27H44N10O6} &\scriptsize \ce{C46H66N12O9} \\
 & \scriptsize 42 (20) &\scriptsize  50 (23) &\scriptsize  68 (37) &\scriptsize  62 (39) &\scriptsize  82 (54) &\scriptsize  123 (73) &\scriptsize  69 (34) &\scriptsize  122 (57) &\scriptsize  130 (63) \\
\hline  \parbox[t]{4mm}{\multirow{6}{*}{\rotatebox[origin=c]{90}{Fragments}}}
 &\scriptsize  \ce{C3H7NO}     &\scriptsize \ce{C4H9NO}      & \scriptsize\ce{C8H9NO}    &\scriptsize$2\times$\ce{C8H6N2} & \scriptsize $2 \times$ \ce{C8H4N2}  & \scriptsize $2\times $\ce{C4H4N2} & \scriptsize\ce{C5H9NO4} &\scriptsize\ce{C12H21N3O3} &\scriptsize\ce{C12H21N3O3} \\
 &\scriptsize   \ce{C4H5NS}    &\scriptsize \ce{C2H5NO}      & \scriptsize\ce{C5H12N2}   &\scriptsize$3\times$\ce{C6H6}   & \scriptsize $3\times$ \ce{C6H6}     & \scriptsize $3\times$ \ce{C6H6}   & \scriptsize\ce{C4H8N2O3} &\scriptsize\ce{C6H9N3O2} &\scriptsize\ce{C11H12N2O2}\\
 &\scriptsize   \ce{C4H6N2}    &\scriptsize$ 2\times$\ce{CH4}  & \scriptsize\ce{C4H4N2}    &\scriptsize\ce{H3N}             & \scriptsize\ce{C8H6O}               & \scriptsize$2\times$ \ce{C7H8}          & \scriptsize\ce{C3H7NO3} &\scriptsize\ce{C6H13NO2} &\scriptsize\ce{C6H14N4O2  \\
 &\scriptsize  \ce{C3H6}       &\scriptsize \ce{C4H4N2}      & \scriptsize \ce{C5H5N}    &                                & \scriptsize\ce{C7H8}                & \scriptsize\ce{C15H15N} &\scriptsize \ce{C5H11NO2} &\scriptsize\ce{C6H13NO2} }&\scriptsize\ce{C6H13NO2}\\
 &                             &\scriptsize \ce{CH5N}        & \scriptsize\ce{C6H6}      &                                &                                     & \scriptsize\ce{C12H9N} &      &\scriptsize\ce{C5H9NO4}&\scriptsize\ce{C4H7NO4} \\
 &                             &\scriptsize \ce{C3H8}       & \scriptsize \ce{CH4}, \ce{H3N}  & & & &                     && \\
\hline
\end{tabular}
\caption{\raggedright
\textbf{Fragment multisets used in this study.} Each column is a different multiset, starting row-wise. Listed are the overall chemical formula of the set, its total number of atoms (heavy atoms in brackets), and the chemical formula of each fragment in the multiset.}
\label{tab:frags}
\end{table*}


\newpage 

\begin{figure}[t]
\centering
\includegraphics[width=0.99\columnwidth]{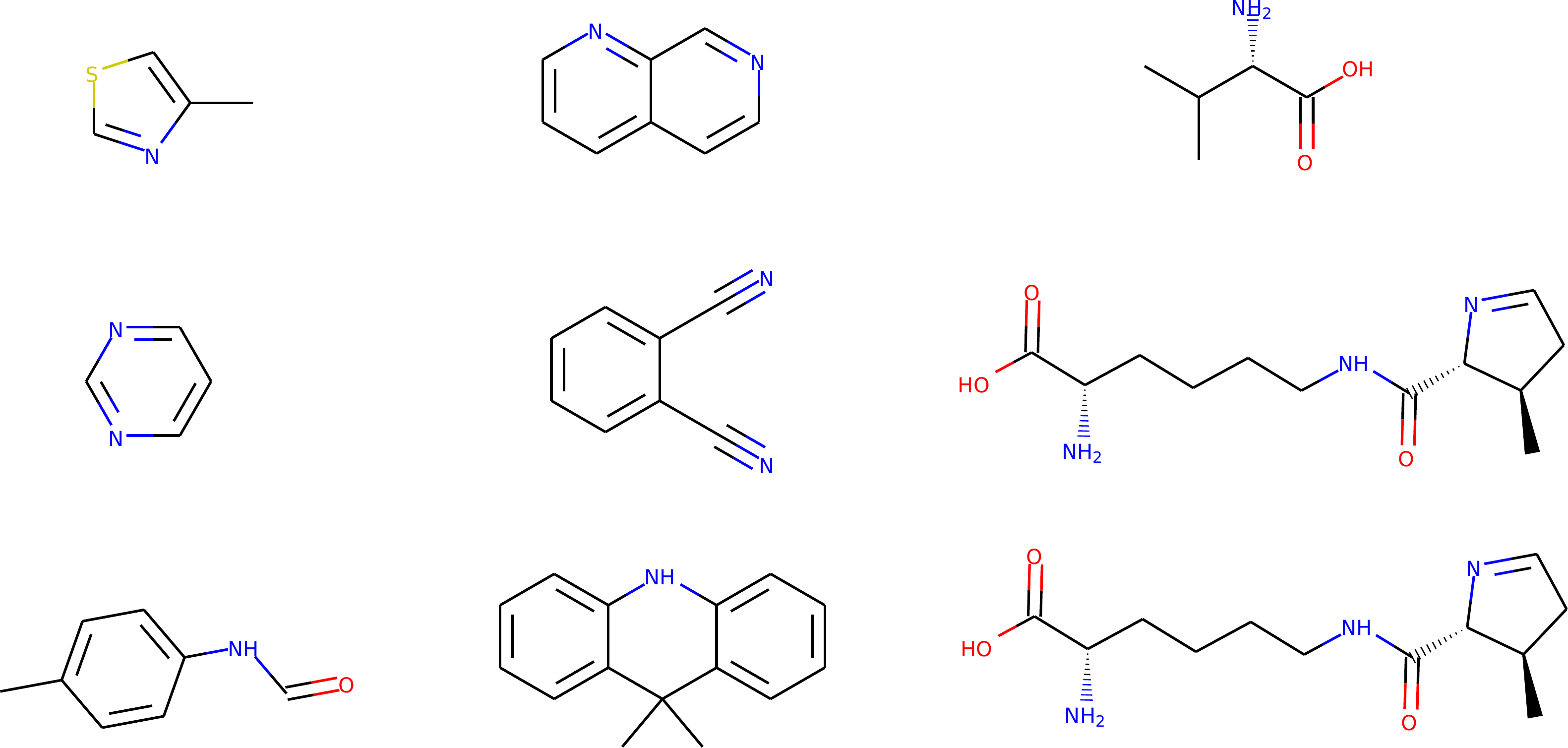}
\caption{\raggedright
Largest fragments in each of the nine multisets. 
Respectively each row is from the drug-like, organic LED, and biomolecule multisets.   
}
\label{fig:lf}
\end{figure}

\begin{figure*}[t]
\centering
\begin{tikzpicture}
    \draw (0, 0) node[inner sep=0] {\includegraphics[width=\textwidth]{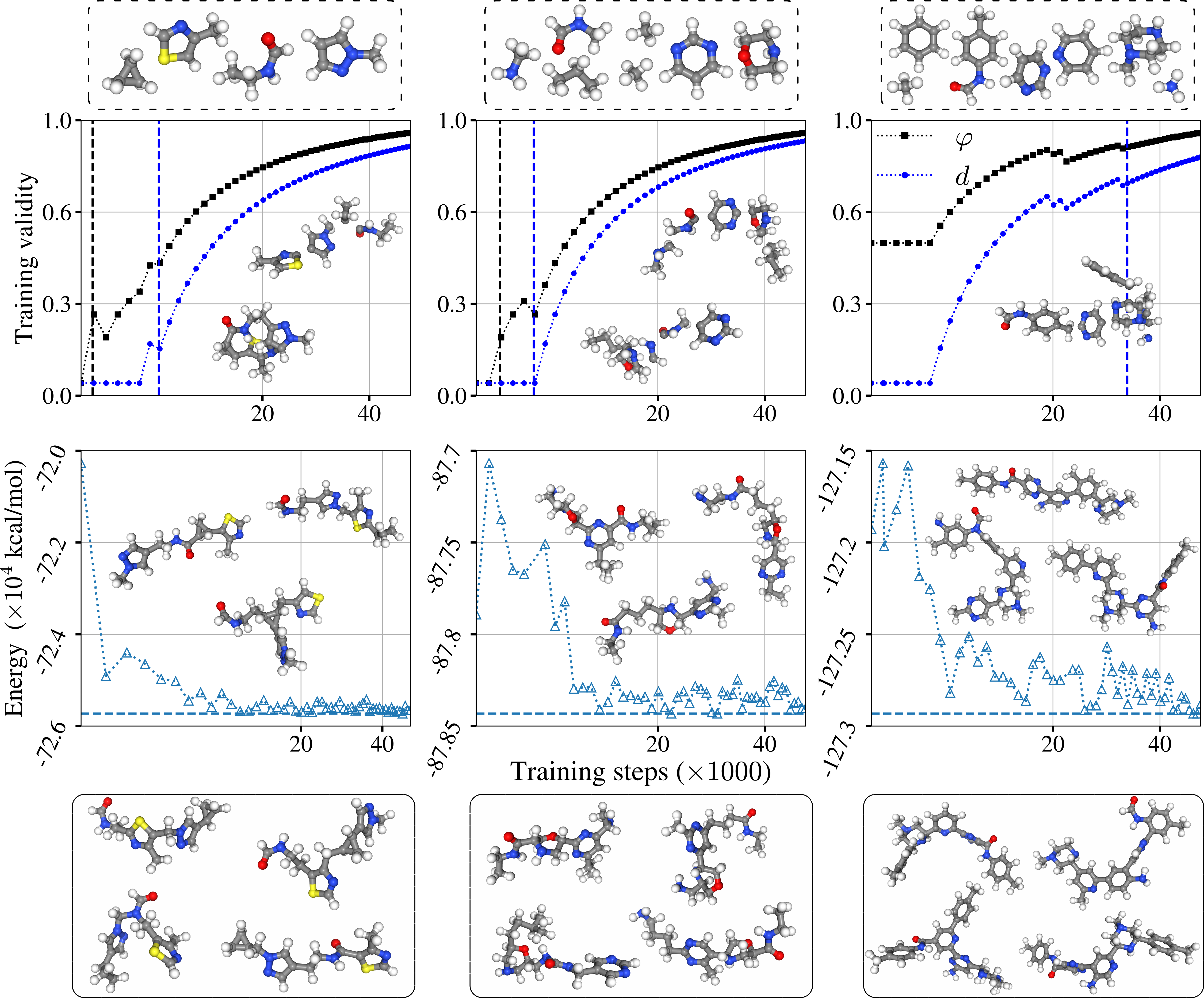}};
    \draw (-8.5, 7.25) node[] {\textbf{a}};
    \draw (-8.5, 5.5) node[] {\textbf{b}};
    \draw (-8.5, 0.75) node[] {\textbf{c}};    
    \draw (-8.5, -4.5) node[] {\textbf{d}};    
\end{tikzpicture}
\caption{\raggedright
\textbf{Drug-like molecules.} Column 1: \ce{C14H22N4OS}, 2: \ce{C16H27N5O2}, and 3: \ce{C29H31N7O}.  
                    \textbf{a.} Three fragment multisets shown in 3D.
                   \textbf{b.} Validity plots for each set with 1-2 examples of invalid molecules. Black lines - rotation  validity ($\varphi$), blue - bond  validity ($d$).
                   \textbf{c.} Energy plots for each set with examples of molecules generated during 10-25K training steps.
                   \textbf{d.} Four examples of molecules produced near the end of training between 30-50K steps.}
\label{fig:drug}
\end{figure*}

\begin{figure*}[t]
\centering
\begin{tikzpicture}
    \draw (0, 0) node[inner sep=0] {\includegraphics[width=\textwidth]{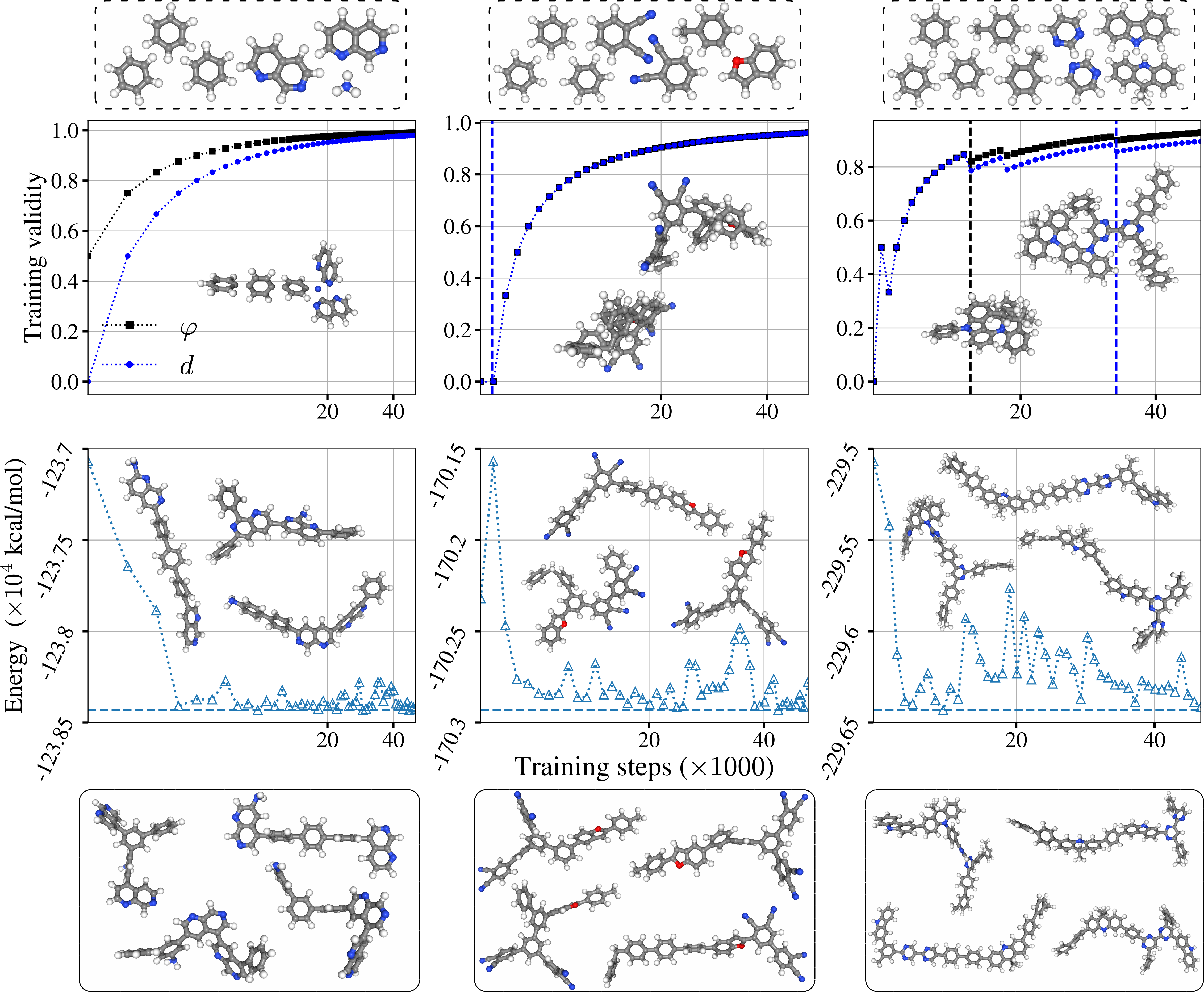}};
    \draw (-8.5, 7.25) node[] {\textbf{a}};
    \draw (-8.5, 5.5) node[] {\textbf{b}};
    \draw (-8.5, 0.75) node[] {\textbf{c}};    
    \draw (-8.5, -4.5) node[] {\textbf{d}};    
\end{tikzpicture}
\caption{\raggedright
\textbf{Organic LEDs.} Column 1: \ce{C34H23N5}, 2:  \ce{C49H28N4O}, and 3: \ce{C67H50N6}.  
                    \textbf{a.} Three fragment multisets shown in 3D.
                   \textbf{b.} Validity plots for each set with 1-2 examples of invalid molecules. Black lines - rotation  validity ($\varphi$), blue - bond  validity ($d$).
                   \textbf{c.} Energy plots for each set with examples of molecules generated during 10-25K training steps.
                   \textbf{d.} Four examples of molecules produced near the end of training between 30-50K steps.
}
\label{fig:oled}
\end{figure*}

\begin{figure*}[t]
\centering
\begin{tikzpicture}
    \draw (0, 0) node[inner sep=0] {\includegraphics[width=\textwidth]{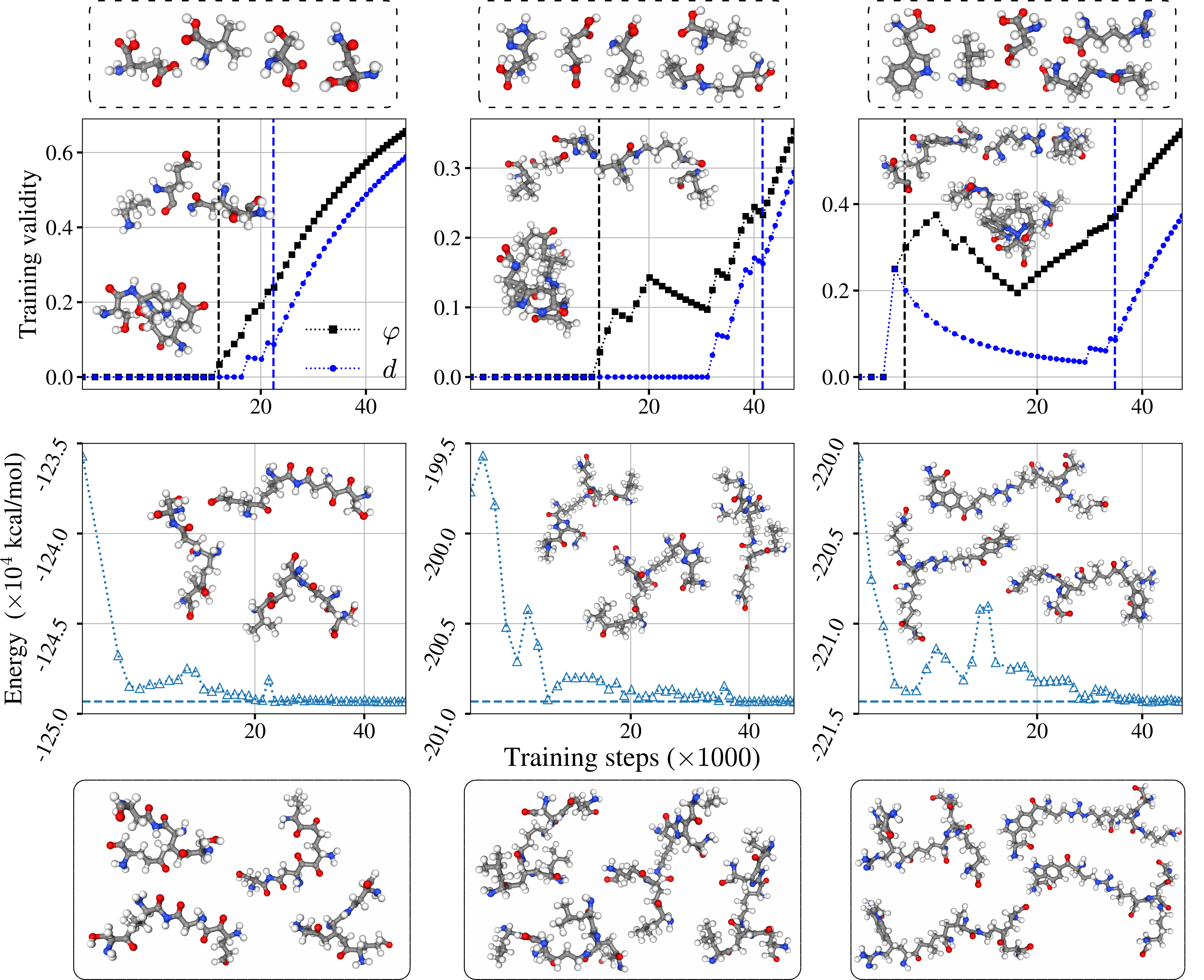}};
    \draw (-8.5, 7.25) node[] {\textbf{a}};
    \draw (-8.5, 5.5) node[] {\textbf{b}};
    \draw (-8.5, 0.75) node[] {\textbf{c}};    
    \draw (-8.5, -4.5) node[] {\textbf{d}};    
\end{tikzpicture}
\caption{\raggedright
\textbf{Biomolecules.} Column 1: \ce{C20H38N6O4}, 2: \ce{C27H44N10O6}, and 3: \ce{C46H66N12O9} 
                    \textbf{a.} Three fragment multisets shown in 3D.
                   \textbf{b.} Validity plots for each set with 1-2 examples of invalid molecules. Black lines - rotation validity  ($\varphi$), blue - bond  validity ($d$).
                   \textbf{c.} Energy plots for each set with examples of molecules generated during 10-25K training steps.
                   \textbf{d.} Four examples of molecules produced near the end of training between 30-50K steps.
}
\label{fig:bio}
\end{figure*}

\begin{figure*}[t]
\centering
\begin{tikzpicture}
    \draw (0, 0) node[inner sep=0] {\includegraphics[width=\textwidth]{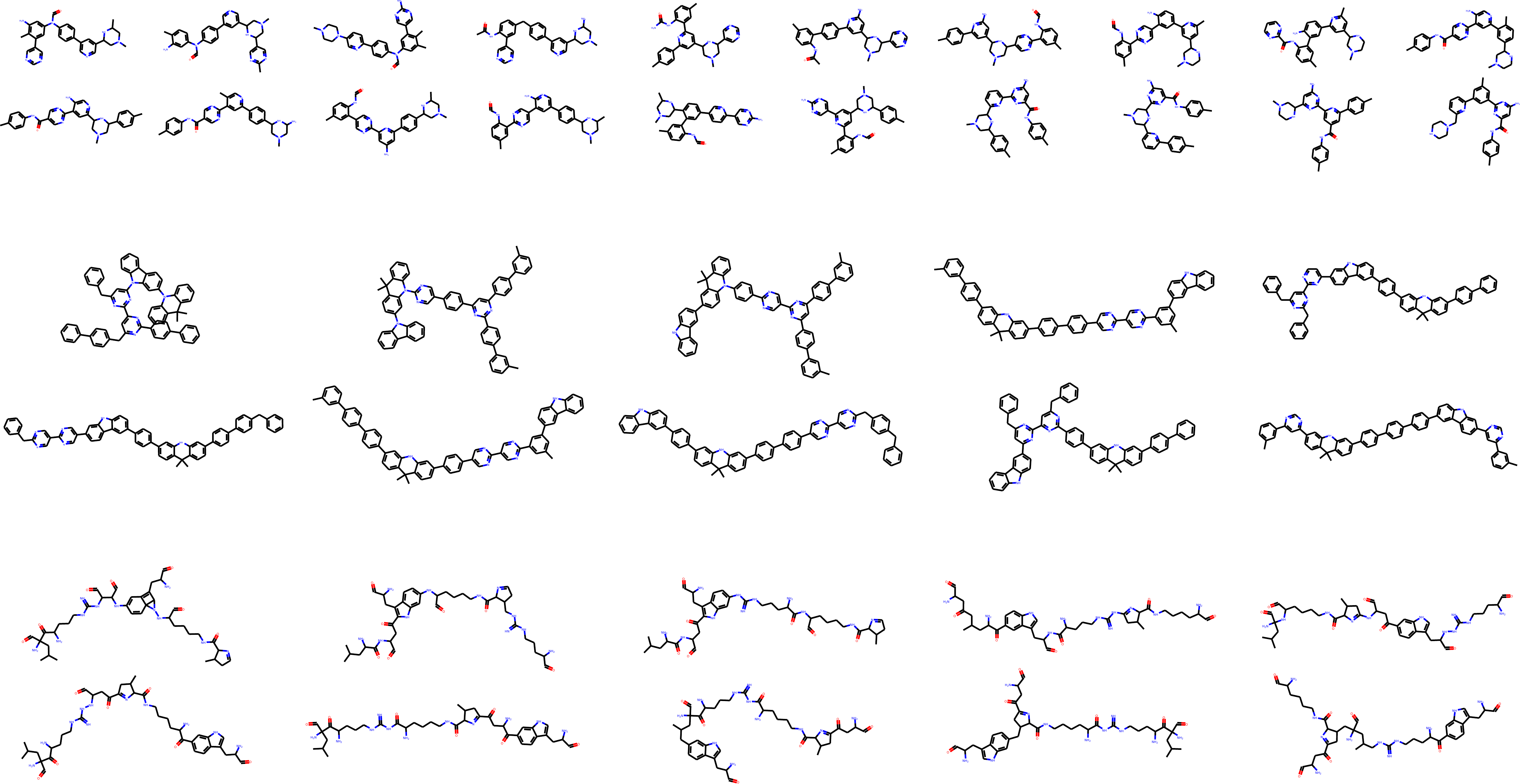}};
    \draw (-8.45, 4.5) node[] {\textbf{a}};
    \draw (-8.35, 1.5) node[] {\textbf{b}};
    \draw (-8.35, -2.0) node[] {\textbf{c}};
\end{tikzpicture}
\caption{\raggedright
\textbf{Discrete Exploration.} Molecular graphs produced using XYZ2MOL from the largest fragment multiset for each distribution. Specifically: 
\textbf{a.} \footnotesize \ce{C29H31N7O},  \textbf{b.} \footnotesize \ce{C67H50N6}, \textbf{c.} \footnotesize \ce{C46H66N12O9}}
\label{fig:des}
\end{figure*}

\section{Results} 

We perform experiments to evaluate the ability of our agent to design valid molecules in Cartesian coordinates, using a variety of fragment types as building blocks.

Specifically, in this work we use our framework to see if our agent can construct molecules using increasingly complex actions in order to build larger more structured molecular distributions. The complexity of the action our agent makes is defined by the size and structure of the molecular fragment being placed. We test our approach on three different kinds of fragment multisets to assess 1) if our agent can design molecules using a variety of small fragments from single atoms to small rings, 
2) if we can build complex molecules using many medium sized fragments of many single rings and fused rings, 
and 3) if we can use even larger actions by combining molecules with several functional groups to build big molecules. Details on the model architecture and hyperparameters are in the Methods section \ref{app:methods}.


\textbf{Molecular distributions.} We apply our framework to a few different molecular distributions that satisfy each type of fragment multiset that 
we wish to test. Each distribution defines a different potential application area for molecular design. 
We train the agent to build molecules using a variety of different fragment sets for distributions of molecules including 
\emph{drug-like} molecules, \emph{organic LEDs} molecules, and \emph{biomolecules}. We use three different fragment multisets for each distribution with increasing size. Table \ref{tab:frags} contains the chemical formula of each set with its total number of atoms as well the formula for every fragment. Figures \ref{fig:drug}\textbf{a},\ref{fig:oled}\textbf{a},\ref{fig:bio}\textbf{a} show the structures of the selected fragments. We also plot the molecular graph of the largest fragment in each multiset in Figure \ref{fig:lf}.


\textbf{Training details.} We train the agent on all nine fragment multisets shown in Table \ref{tab:frags} for 50K (thousand) steps. Since we are opting for a general demonstration of our framework for the prior geometry, we use fixed coordinates for each fragment that do not change during training.

\textbf{Offline Evaluation.} We generate molecular structures at specific evaluation points -- terminal states during training at every 1K steps starting from step 100. We evaluate the molecular structures in order to determine how well the agent learns. To this end, we analyze the generated molecular structures' validity and energy.  

\textbf{Validity.} To determine whether a generated molecular structure is considered valid we use the XYZ2MOL \cite{xyz2mol} tool based on \cite{kim2015universal}. 
If XYZ2MOL can successfully convert the structure into a molecular graph and produce a valid SMILES string then the structure is considered valid.
This is further divided into bond validity and rotation validity. For rotation validity, we allow the smiles string to contain '.' -- this happens when the agent has placed the fragments too far from the anchor locations and the smiles string includes unconnected fragments floating in space. 
For bond validity, we classify smiles strings with periods as invalid. We plot the cumulative valid ratio (from step 100) at each evaluation point -- that is the ratio of valid to total molecules generated from step 100 to the current evaluation point. We plot both forms of validity on the same plot -- bond validity in blue and rotation validity in black. Furthermore, we plot the time where the agent first begins to generate molecules with rotation validity using a black dashed vertical line and we plot the time where the agent begins to generate molecules that are always completely valid using a blue dashed vertical line. On each plot, we show from the first five evaluation points ($<$ 5K steps) one example of a molecular structure that is completely invalid and one that is rotation valid.

\textbf{Energy.} To assess whether or not the agent does in fact learn to produce lower energy molecules we plot the energy in kcal/mol over the entire training regime of 50K steps. We use the molecules produced from the evaluation points during training and calculate their energy with the same method used in the reward. We plot the lowest energy reached as a dashed horizontal line. For the plots, we omit the energy from molecules generated at step 100 if it is significantly larger than the subsequent energies. In addition, we display a few examples of valid molecular structures obtained throughout training from two regions: 1) on the energy plots, we plot structures from the time where the agent first starts to generate valid structures and is only beginning to generate lower energy molecules (typically $<$ 25K steps). 2) We plot separately the molecules obtained from the time when the agent has learned to build lower energy structures (typically $>$ 25K steps). 

\textbf{Drug-like molecules.} We form three fragment multisets by breaking molecules with high QED \cite{bickerton2012quantifying} at rotatable bonds. 
Two molecules with chemical formulas \ce{C14H22N4OS} and \ce{C16H27N5O2} were taken from the ZINC15 database and the third one is Imatinib (\ce{C29H31N7O}) -- an oral chemotherapy medication used to treat chronic myelogenous leukemia and some other types of cancer. These fragment sets include a variety of structures from small molecules with one heavy atom to larger rings with branches. \ce{C14H22N4OS} has 2 five-membered rings, a cyclopropane fragment, and a 5-atom linear structure. \ce{C16H27N5O2} has 2 singular carbons in the form of \ce{CH4}, 3 linear structures with 2-4 heavy atoms, and 2 rings. \ce{C29H31N7O} has 5 rings -- 2 of them with branches -- and singular carbon and nitrogen (\ce{CH4} and \ce{NH3}).

The results from training the agent on this fragment set are shown in Figure \ref{fig:drug}. For column/set 1, the agent takes around 10K steps to generate valid, low energy molecules. For column/set 2, the agent takes around 15K steps to perform the same task.     
For column/set 3, the agent takes longer -- around 35K steps to generate valid, low energy molecules. For this set, convergence is noisy but the structures generated throughout look quite reasonable (Figure \ref{fig:drug}\textbf{d}).   
 
\textbf{Organic LEDs.} The emissive layers of organic light-emitting diode (OLED) devices are made from electroluminescent molecules. Due to their high efficiency and superior color properties, OLEDs are being widely used in manufacturing displays for digital electronics (TVs, smartphones, etc.). To build some fragment multisets for OLED molecules, we took a few fragments from a library used in high-throughput virtual screening for OLED design \cite{gomez2016design}. The resulting multisets have the following chemical formulas:
\ce{C34H23N5}, \ce{C49H28N4O}, \ce{C67H50N6}. \ce{C34H23N5} has 3 benzene rings, a nitrogen (\ce{NH3}), and 2 fused bicyclic structures. \ce{C49H28N4O} has 3 unsubstituted benzene rings, 1 \ce{CH3}-substituted benzene (toluene), a bicyclic fused ring, and 2 benzene rings with 2 nitrile groups. \ce{C67H50N6} has 3 benzenes, 2 toluenes, 2 tricyclic fused structures, and 2 aromatic heterocycles.

The results from training the agent on these multisets are shown in Figure \ref{fig:oled}. For column/set 1 and 2, the agent takes around 5-10K steps to generate molecules with lower energy but learns to generate valid molecules very quickly -- within the first 2K steps.     
For column/set 3, the agent takes 5K steps to learn to generate valid molecules. The energy distribution of the generated molecules is more noisy and achieves consistent convergence to low energy structures at 35K steps. For these OLED fragment sets, the produced molecules in \ref{fig:oled}\textbf{c,d} do resemble typical OLED-like molecules \cite{gomez2016design}. In particular, column one exhibits molecules with interesting 2D and 3D symmetries, which is remarkable given that the reward is only based on energy.   

\textbf{Biomolecules.} To test if we can use very large actions, we use amino acids as fragments in order to build molecules. We do not restrict the possible connectivities and allow any hydrogen to be used as an anchor point when adding the amino acids to the molecules. Thus, the explored space is not limited to peptides but instead represents a broader range of natural product-like biomolecules.
The difficulty for this case arises because the fragments have many hydrogens and therefore more locations to use as an anchor when performing an action. We experiment with three different sets of amino acids: 1) \{E,N,S,V\} 2) \{H,E,L,L,O\} 3) \{W,O,R,L,D\} (see \ref{fig:bio}\textbf{a}).

The results from training the agent on these amino acid multisets are shown in Figure \ref{fig:bio}. For column/set 1, the agent takes around 20K steps to generate completely valid molecules with consistently lower energy. For the multisets with 5 amino acids, the agent takes substantially longer but eventually learns to generate valid molecules at around 40K steps. The agent takes 10K steps to initially generate molecules with lower energy but only after 20K steps does it really start to converge. For these amino acid actions, the produced molecules in \ref{fig:bio}\textbf{c,d} resemble the 3D geometry of peptides and have similar structure: many branches extending from a longer backbone. 

\textbf{Discrete Exploration.} The agent explores the discrete space of possible attachments and the continuous space of possible conformations. Discrete space is explored when the agent selects hydrogens which are to be replaced with fragments. To demonstrate the exploration of the discrete space, we plot 10 or 20 molecules as graphs produced from XYZ2MOL for each of the largest fragment multiset in all three distributions (Figure \ref{fig:des}).

\textbf{Comparison to other approaches.} Previous approaches generate molecules by connecting fragments with bonds, while our agent has to learn to exactly position fragments in Cartesian coordinates -- this makes exploration more challenging. Regardless, our agent can learn comparable validity and after training will produce molecules that are consistently valid ($> 95\%$) despite the added difficulty.
For MolGym, which uses a fixed atom multiset as well as an energy based reward, we cannot make a direct comparison since they place single atoms, however in general our approach consistently generates valid structures within 50k steps, whereas MolGym takes 2-5 times longer to generate with a much lower validity. In addition, Molgym is only capable of building small molecules with less than 30 atoms.

\textbf{Limitations} We discuss some limitations of this approach and how to overcome them.

\emph{Fixed Space.} Using a fixed multiset of fragments is a double edged sword -- despite its convenience, we are restricted to design molecules that have specific formulas and sets of substructures. However, one could also pair the approach with a pretrained generative model to suggest substructures and fine-tune it for a particular task \cite{chen2021fragment}. 

\emph{Reward Issues.} Similar to other approaches that use energy as a reward, the agent will have issues with low molecular diversity after convergence. To address this, the reward can be combined with other metrics such as drug-likeness \cite{bickerton2012quantifying} or synthesizability \cite{ertl2009estimation}.

\section{Potential Applications}

Our approach is versatile and scalable -- it can be modified for many potential applications such as small molecule and peptide drug design, as well as design of ligands and organocatalysts. Modifying the reward function will make the agent optimize a different property other than total energy. For example, rewarding hydrophilicity instead of just energy can make the model assemble different functional groups from the same fragments. Since the system operates in three dimensions, imposing geometric constraints in the environment will make the agent learn to assemble molecules that directly optimize some desired geometric properties. For example, the environment can be constrained to the shape of the binding pocket and the agent can be rewarded for a good docking score and stability of the molecule. The environment can also be modified to be more dynamic, which will make the learning process more robust and will focus the agent on producing the best binders. Similarly, ligands for metal catalysts can be constrained to a desired shape and rewarded for their electronic properties.

\section{Conclusion}

This work is a step towards flexible molecular design in 3D space. By using fragments with predefined shapes, we place an implicit prior on the geometry that the agent can use to build significantly more complex molecular structures. This approach aligns with the intuition of an organic chemist who, when faced with a task to modify a molecule, would think of it in terms of adding new fragments and functional groups rather than single atoms.

We have described a novel RL formulation for molecular design in 3D space that is guided by quantum mechanics. We proposed a hierarchical architecture for our agent that uses high-level actions -- placing whole molecular substructures -- in order to build a molecule. We have demonstrated that our model can efficiently build a range of molecules from different distributions using a variety of fragments that range in size and structure. Furthermore, we have shown that our approach is scalable, as our agent can build molecules with more than 100 atoms.


\section*{Acknowledgements}
 A.A.-G. acknowledges funding from Dr. Anders G. Fr{\o}seth. A.A.-G. also acknowledges support from the Canada 150 Research Chairs Program, the Canada Industrial Research Chair Program, and from Google, Inc.
Models were trained using the Canada Computing Systems \cite{baldwin2012compute}.






\bibliographystyle{apsrev4-1}
\bibliography{references}


\clearpage

\appendix

\section{Methods} \label{app:methods}

\subsection{Quantum-Chemical Calculations}\label{app:quantum}

For the calculation of the energy $E$, we use the fast semi-empirical Parametrized Method 6 (PM6) \citep{Stewart2007} using the software package \textsc{Sparrow} \citep{Husch2018a,Bosia2019}.
For each calculation, a molecular charge of zero and the lowest possible spin multiplicity are chosen.
All calculations are spin-unrestricted.  

For the quantum-chemical calculations to converge reliably, we ensured that atoms are not placed too close ($<$ 0.6~\AA) nor too far away from each other ($>$ 2.0~\AA).
If the agent places an atom outside these boundaries, the minimum reward of $-10$ is awarded and the episode terminates.

\subsection{Policy Gradient Details}

We employ PPO \citep{schulman2017proximal} to learn the parameters $(\theta, \phi)$ of the actor $\pi_\theta$ and critic $V_\phi$ as it can handle both continuous and discrete action spaces. 
To help maintain sufficient exploration throughout learning, we include an entropy regularization term over the policy's categorical distributions.
 
Policy gradient methods learn a parametrized policy $\pi_\theta$ by performing gradient ascent in order to maximize $J(\theta)$.
The most widely used gradient estimator is the REINFORCE policy gradient $\nabla_\theta J(\theta) = \mathbb{E}[\nabla_\theta \pi(a_t \vert s_t) \hat{A}_t]$ \citep{williams1992simple}, where $\hat{A}_t$ is an estimator of the advantage function.
A recent improved extension, proximal policy optimization (PPO) \citep{schulman2017proximal} employs a clipped surrogate objective that constrains the policy updates in order to improve the stability during learning.
Denoting the probability ratio between the updated and the old policy as $r_t(\theta) = \pi_\theta(a_t \vert s_t) / \pi_{\theta_{\text{old}}}(a_t \vert s_t)$,
the clipped objective $J^\text{CL}$ is given by
\begin{equation*}
    J^\text{CL}(\theta) = \mathbb{E} \left[ \min(r_t(\theta)\hat{A}_t, \text{clip}(r_t(\theta), 1-\epsilon, 1+\epsilon)\hat{A}_t) \right],
\end{equation*}
where $\hat{A}_t$ is an estimator of the advantage function and $\epsilon$ is a hyperparameter that controls the interval beyond which $r(\theta)$ gets clipped.
To further reduce the variance of the gradient estimator we use actor-critic methods \citep{konda2000actor}. 
If the actor and critic share parameters, the objective becomes
\begin{equation*}
    J^\text{AC}(\theta) = \mathbb{E} \left[J^\text{CL}(\theta) - c_1 J^\text{V} + c_2 \mathbb{H}[\pi_\theta \vert s_t] \right],
\end{equation*}
where $c_1$, $c_2$ are coefficients, $J^\text{V} = (V_\theta(s_t) - V^\text{target})^2$ is a squared-error loss, and $\mathbb{H}$ is an entropy regularization term to encourage sufficient exploration.
 Further, PPO employs a generalized advantage estimator (GAE) with
\begin{equation*}
    \hat{A}_t = \Delta_t + (\gamma\lambda) \Delta_{t+1} + \cdot\cdot\cdot + (\gamma\lambda)^{T-t+1} \Delta_{T-1},
\end{equation*}
where $\Delta_t = r(s_t, a_t) + \gamma V^\pi(s_{t+1}) - V^\pi(s_t)$ is the temporal-difference (TD) error and $\lambda$ is a hyperparameter.

\subsection{The Critic}

The critic needs to compute a value for the entire state $s$.
A pooling operation is necessary to aggregate over all atom embeddings added in the molecule with every additional fragment that is placed. Hence, we compute the sum over all molecular atom embeddings $\B h_v $ so the critic is given by
\begin{equation}
    V_\phi(s) = \text{MLP}_\phi \left(\sum_{v\in \mathcal{M}}  \B h_v \oplus \B h_{\scriptscriptstyle \mathcal{F}} \right),
\end{equation}
where $\text{MLP}_\phi$ is an MLP that computes value $V$.


\subsection{Predicting the sign of the Rotation Angle} \quad
The rotation angle $\varphi $ for the attached fragment is difficult to learn because we choose the nearest neighbours from the attachment atoms to calculate this dihedral.
We can overcome this issue by just learning the absolute value $|\varphi| \in [0, \pi]$ instead as well as the sign of $\varphi$, $\text{sgn}(\varphi) \in \{+1, -1\}$. So that $p(\text{sgn}(\varphi) \vert \, \vert\varphi\vert ,u_f, f, v_{\scriptscriptstyle\mathcal{M}}, s ) = \text{Ber}(p_{\text{sgn}(\varphi)})$
which is produced using from a MLP using a sigmoid activation $p_{\text{sgn}(\varphi)}) =\textsc{MLP}[\B h_{u_f} \oplus \B h _{v_{\scriptscriptstyle\mathcal{M}}} \oplus \B h_{\scriptscriptstyle\mathcal{F}} \oplus \bm x_f ]$.

\subsection{Model Architecture}

We initialize the biases of each MLP with $0$ and each weight matrix as a (semi-)orthogonal matrix. After each hidden layer, a ReLU non-linearity is used.
Further, we rescale the means of the continuous actions predicted so that the fragment distance mean is within the max/min distance range and the mean rotation angle predicted is between 0 and $\pi$.


\subsection{Hyperparameters}

The hyperparameters are as follows.
The max/min fragment distance range in (\AA) is $[1.10, 2.10]$. During training we use 8 Workers.
Clipping $\epsilon=0.2$. Gradient clipping : $0.5$.
GAE parameter $\lambda=0.97$.
VF coefficient $c_1=0.5$.
Entropy coefficient $c_2=0.01$.
Training epochs: $5$.
Adam stepsize: $3 \times 10^{-4}$.
Discount $\gamma$:  $1$.
Minibatch size : $100$.
Hidden layer size in MLPs: $128$.

The hyperparameters for SchNet \citep{Schutt2017} used in all experiments are
interactions: 3, 
Cutoff distance (\AA): $5.0$, 
Number of filters: $128$, 
Number of atomic basis functions: $64$.




    
    
    
     
     

     
     
     
     
     






\end{document}